\documentclass[conference]{IEEEtran}
\IEEEoverridecommandlockouts
\usepackage{cite}
\usepackage{amsmath,amssymb,amsfonts}
\usepackage{algorithmic}
\usepackage{graphicx}
\usepackage{textcomp}
\usepackage{xcolor}
\def\BibTeX{{\rm B\kern-.05em{\sc i\kern-.025em b}\kern-.08em
    T\kern-.1667em\lower.7ex\hbox{E}\kern-.125emX}}
\begin{document}

\title{ASAP: Accurate semantic segmentation for real time performance\\

}

\author{\IEEEauthorblockN{Jae Hyun Park}
\IEEEauthorblockA{\textit{AI tech team} \\
\textit{Lotte Data Communication Company}\\
Seoul, South Korea \\
jaehyun-park@lotte.net}
\and
\IEEEauthorblockN{Su Bin Lee}
\IEEEauthorblockA{\textit{AI tech team} \\
\textit{Lotte Data Communication Company}\\
Seoul, South Korea \\
leesubin@lotte.net}
\and
\IEEEauthorblockN{Eon Kim}
\IEEEauthorblockA{\textit{AI tech team} \\
\textit{Lotte Data Communication Company}\\
Seoul, South Korea \\
eon.kim@lotte.net}
\and
\IEEEauthorblockN{Byeong Jun Moon}
\IEEEauthorblockA{\textit{AI tech team} \\
\textit{Lotte Data Communication Company}\\
Seoul, South Korea \\
bj\_moon@lotte.net}
\and
\IEEEauthorblockN{Da Been Yu}
\IEEEauthorblockA{\textit{AI tech team} \\
\textit{Lotte Data Communication Company}\\
Seoul, South Korea \\
db.yu@lotte.net}
\and
\IEEEauthorblockN{Yeon Seung Yu}
\IEEEauthorblockA{\textit{AI tech team} \\
\textit{Lotte Data Communication Company}\\
Seoul, South Korea \\
yys4000@lotte.net}
\and
\IEEEauthorblockN{Jung Hwan Kim}
\IEEEauthorblockA{\textit{AI tech team} \\
\textit{Lotte Data Communication Company}\\
Seoul, South Korea \\
jhwan\_kim@lotte.net}
}

\maketitle

\begin{abstract}
Feature fusion modules from encoder and self-attention module have been adopted in semantic segmentation. However, the computation of these modules is costly and has operational limitations in real-time environments. In addition, segmentation performance is limited in autonomous driving environments with a lot of contextual information perpendicular to the road surface, such as people, buildings, and general objects. In this paper, we propose an efficient feature fusion method, Feature Fusion with Different Norms (FFDN) that utilizes rich global context of multi-level scale and vertical pooling module before self-attention that preserves most contextual information while reducing the complexity of global context encoding in the vertical direction. By doing this, we could handle the properties of representation in global space and reduce additional computational cost. In addition, we analyze low performance in challenging cases including small and vertically featured objects. We achieve the mean Interaction of-union(mIoU) of 73.1 and the Frame Per Second(FPS) of 191, which are comparable results with state-of-the-arts on Cityscapes test datasets.
\end{abstract}


\section{Introduction}
Semantic segmentation is a per-pixel classification which predicts pixel by pixel. Including biomedical and human-machine interaction, semantic segmentation has been widely researched \cite{kim2019u, li2020humans}.

In particular, segmentation used in autonomous driving, such as depth estimation and free space, operates in real time and requires fast inference speed and high performance.
To improve inference speed, aligned feature maps at adjacent levels used to balance performance and inference speed in segmentation task \cite{li2020semantic}. ladder-style lightweight decoder is designed for upsampling low spatial resolution \cite{wang2021swiftnet}.

To achieve high accuracy, segmentation models require global contextual information and capabilities with multi-level semantics. Some studies include a self-attention module, which helps to concentrate contextual features \cite{zhao2018icnet} to satisfy accuracy. Other studies propose the feature fusion module, which combine multi-level features \cite{fan2021rethinking, yu2021bisenet}. However, these modules, which contain convolution-based operations to fusion multi-level features, require huge computational complexity and memory storage. 

\begin{figure}[htb!]
\centering{\includegraphics[width=85mm]{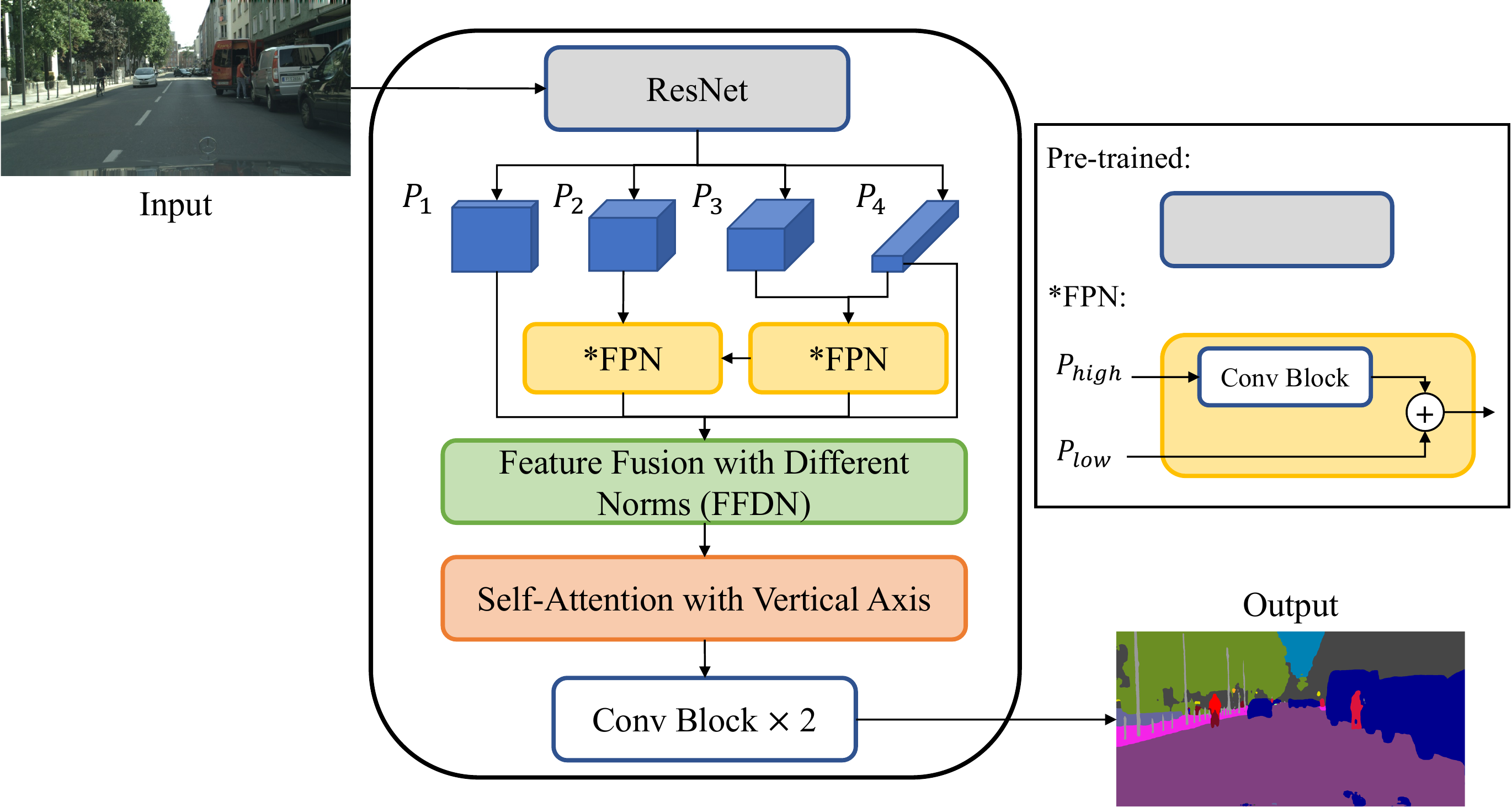}}
\caption{The overview of our proposed fusion module. We aggregate features with different resolution by using normalizations and focus on vertically directional objects by using self attention with vertical axis. *FPN is simply modified version of neck of FPN. Conv Block includes convolution with 3$\times$3 and batch normalization and ReLU activation.}
\label{overview}
\end{figure}

In order to reduce the amount of computation while not dropping the accuracy, we attempt to exploit normalization technics in feature fusion of semantic segmentation. In U-GAT-IT \cite{kim2019u}, spatial and semantic contents are considered adequately by using adaptive normalizations [12, 13] to reflect image content such as style and geometry information. Inspired by these approaches, we propose an efficient Feature Fusion with Different Norms (FFDN) where layer normalization and instance normalization are used in aggregating features of different layers as shown in Fig 2. These normalization methods allow the segmentation model to obtain exact object location from spatial information and detailed parts of object from content information with low computational complexity. FFDN receives multi-level features obtained from simply modified FPN (*FPN) as input and combines them to capture global properties of representations. The *FPN is shown in Fig \ref{overview}.


One of the challenging problems with semantic segmentation is considering specific directions, such as vertical or diagonal (e.g., people, pole). Since a general convolution or pooling operation uses uniform kernels with the same height and width, they do not consider context consistency in a specific direction and thus fail to cover vertical objects.

Striping Pooling \cite{hou2020strip} analyzes the directional consistency and
uses a vertical pooling operation, but requires high computational complexity caused by additional parameters in convolution modules.

\begin{figure}[htb!]
\centering{\includegraphics[width=75mm]{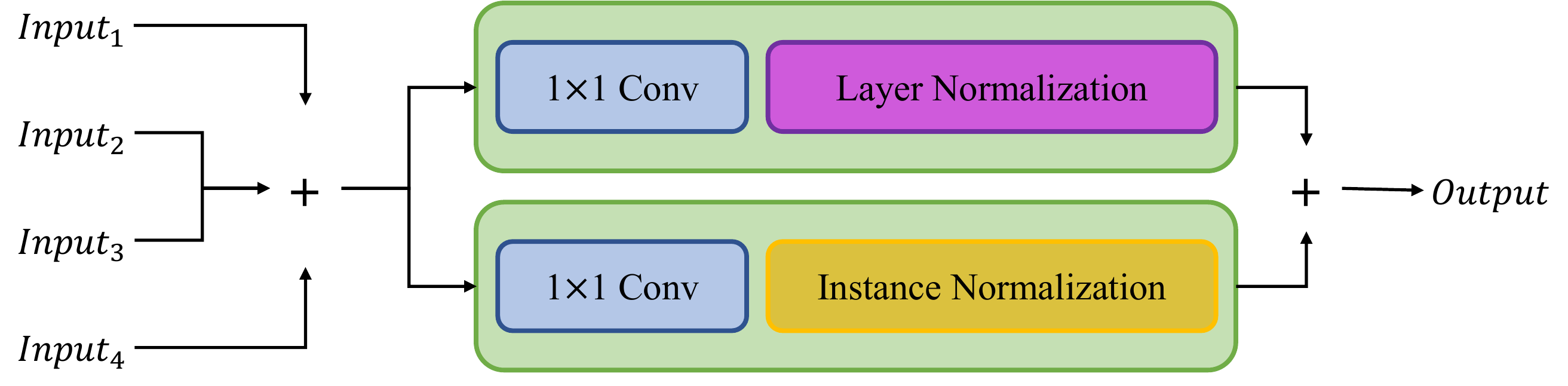}}
\caption{The module of Feature Fusion with Different Norms in Figure 1. The module consists of 1$\times$ 1 convolution layer and different normalizations respectively.}
\label{ffu}
\end{figure}

We propose novel attention module, which could achieve directional consistency by striding vertical direction in self-attention. As self-attention is used to capture global perspective features, using a vertical pooling module in the conventional self-attention method can reduce the size of attention maps and enhance directional consistency. In addition, we use skip-connection to connect the input feature and output of the attention module to consider the directional context information in the horizontal direction as well. In other words, the use of vertical pooling can reduce computational costs by creating lightweight self-attention maps while aggregating global context information along the vertical axis.

In this paper, we cover comparable accuracy and fast inference by 1) lightweight FFDN capturing spatial and semantic properties by using normalization methods and 2) by performing vertical pooling in self-attention to enhancement directional consistency and utilize light attention with low computational cost.
The experimental results show that the proposed ASAP(\textbf{A}ccurate \textbf{S}emantic segment\textbf{a}tion for real time \textbf{P}erformance) model has a higher performance of 73.1mIoU/191FPS in Cityscapes \cite{cordts2016cityscapes} test datasets than conventional methods.

\section{Proposed method}
We propose two main contributions of this paper in this section. Both of contributions target high performance and fast inference speed. 

First, we design new feature fusion module. From the pre-trained network, features in different levels have different properties. feature fusion approaches have been studied to deal with these different properties. \cite{zhao2018icnet, yu2021bisenet}. 

For exploiting different properties of multi level features, We propose a Feature Fusion with Different Norms (FFDN) where layer normalization and instance normalization are used in aggregating different features as shown in Fig \ref{ffu}. Multi-level features are combined using *FPN (a simple modification of the Feature Pyramid Network \cite{lin2017feature}), and the result becomes the input to the FFDN as shown in Fig \ref{overview}.


Fig \ref{overview} shows the overview of our proposed fusion module. $P_{i}$ represents feature level with resolution of $1/2^{i+1}$ in $i$th convolution block. If input resolution is 1024$\times$1024, 7th feature level resolution is 128$\times$128. We use $P_{1} \sim P_{4}$ and scale fused features to the shape of $P_{1}$, because $P_{1}$ has rich information of the input as explained in \cite{vu2022hybridnets, tan2020efficientdet}. 

In combining all features, layer normalization \cite{ba2016layer} and instance normalization \cite{ulyanov2016instance} are introduced into the feature fusion module to consider properties of multi level features

Layer normalization normalizes summed output features from a convolution layer. 

\begin{equation}
\begin{split}
\mu^{l}=\frac{1}{H}\sum^{H}_{i=1}a^{l}_{i}\hspace{3em}\quad\\
    \sigma^{l} = \sqrt{\frac{1}{H}\sum^{H}_{i=1}(a^{l}_{i}-\mu^{l})^{2}},
\end{split}
\end{equation}
where H denotes the number of hidden units in a layer. Under the layer normalization, all the hidden units in a layer share the same normalization terms $\mu$ and $\sigma$. Through the layer normalization, output features preserve each properties of different convolution layers. 

An Instance normalization normalizes specific instance in a convolution layer. 
\begin{equation}
\begin{split}
y_{itjk}=\frac{x_{tijk}-\mu_{ti}}{\sqrt{\sigma}^{2}_{ti}+\epsilon},\hspace{6em}\quad\\
    \mu_{ti} = \frac{1}{HW}\sum^{W}_{l=1}\sum^{H}_{m=1}x_{tilm},\hspace{3em}\quad\\
    \sigma^{2}_{ti}=\frac{1}{HW}\sum^{W}_{l=1}\sum^{H}_{m=1}(x_{tilm}-mu_{ti})^{2}.
\end{split}
\end{equation}

Using FFDN has two contributions. One is that FFDN could replace the global representations which guides contextual information from multi level features with simply fusion multi level features with normalizations considering spatial and semantic properties.
In an image-to-image translation method \cite{kim2019u}, the AdaLIN, which balances layer normalization and instance normalization, is proposed to capture multiple properties of features in global perspective. In \cite{kim2019u}, layer normalization could capture spatial information while instance normalization could capture content information e.g., style in features. Likewise image-to-image translation, semantic segmentation also could achieve exact object location from spatial information and detailed parts of object from content information by using both of two normalizations.

The other is that FFDN could reduce computational complexity. In general, some fusion approaches \cite{huang2019ccnet,fu2019dual, yuan2018ocnet} adopt multiple convolution/pooling operations to generate the global feature in aggregating multi level features, but those operations require huge computational complexity and additional memory storage. 
Our FFDN contains layer/instance normalizations, whose complexities are $O(C_{output}\times H_{output}\times W_{output})$ in calculating $\mu, \epsilon$, and one simple 1x1 convolution layer for channel reduction while the complexity of convolution layer is $O(N\times K\times K\times C_{input}\times C_{input}\times H_{output}\times W_{output})$ where $N$ is the number of convolution operations and $K$ is the kernel size.
Therefore, FFDN achieves efficient feature fusion bridging multi level features adaptively without any additional parameters.

\begin{figure}[h]
\centering{\includegraphics[width=75mm]{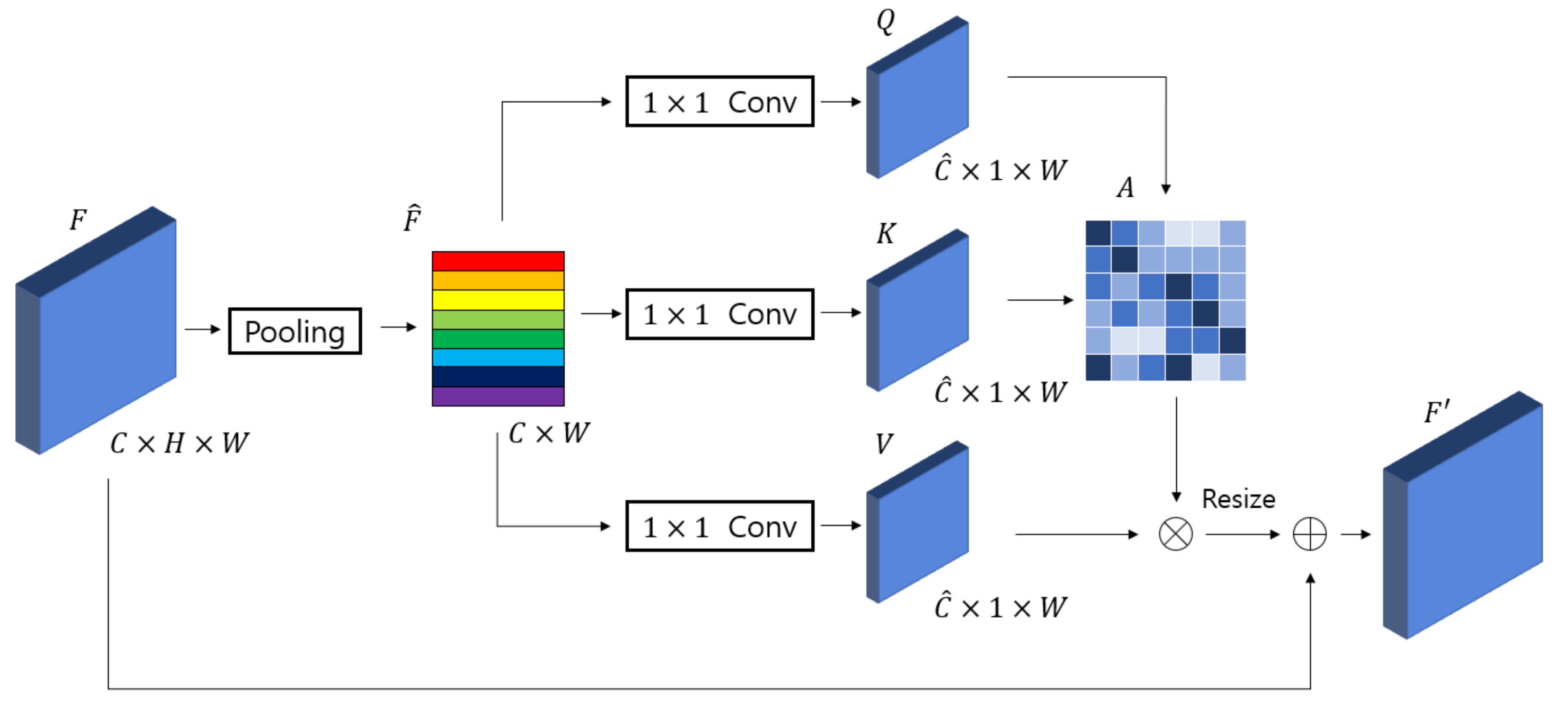}}
\caption{Illustration of the self-attention with vertical axis.}
\label{vertical}
\end{figure}

Second, we design a new self-attention with directional consistency. Self-attention module has been a key component in modern networks, which captures long-range contextual information. However, the general attention module could not have directional consistency since pooling layer operates based on a uniform kernel. The attention has limitation on discriminating shallow objects e.g., person, pole which are characterized by vertical direction. In addition, the attention requires quadratically computation increases depending on input feature size. Therefore, we propose the vertical pooling operation in advance before forwarding self-attention module. This operation not only aggregates contextual information along the vertical axis, but takes advantage of lightweight attention with low computational cost. We explain our proposed self-attention with directional consistency below.

Given an intermediate feature $F \in \mathbb{R} ^{C \times H \times W}$ with $C$ channels, we downsample $F$ using the average pooling layer of $H \times 1$ to focus mainly vertical features $\hat F \in \mathbb{R} ^{C \times 1 \times W}$. $Q$ and $K$ are generated by $1\times 1$ convolution, resulting in $\{Q, K\} \in 
\mathbb{R}^{\hat C \times 1 \times W}$. Based on $Q$ and $K$, general attention map is calculated as follows:
\begin{equation}
A_{j,i} = \frac{exp(Q_{j}\cdot K_{i})}{\Sigma^{W}_{i=1}exp(Q_{j}\cdot K_{i})},
\end{equation}
where $i,j$ indicate each index of $Q, K$ respectively.
$A_{j,i}$ enforces the representation $V \in \mathbb{R} ^{\hat C \times 1 \times W}$ along with vertical axis by using Equation \ref{eq:att}. We perform an element-wise summation with $F$ to obtain the final output $F^{\prime} \in \mathbb{R} ^{\hat C \times H \times W}$ 

\begin{equation}
F'_{j} = R(\sum^{W}_{i=1}A_{j,i}\cdot V_{i}) + F_{j},
\label{eq:att}
\end{equation}
where $R(\cdot)$ refers to resize function.

In challenging cases including small and vertical featured objects like pole and fence, our self-attention with vertical pooling could focus on vertical important representation in perspective of global features. For tracking horizontal feature, we use skip connection with input feature $F_{j}$. 
In addition, the computational cost of our attention module is $O(W^2)$ while the computational cost of general self-attention is $O((HW)^2)$, which increases quadratically corresponding the size of inputs. Therefore, our attention module not only reduces computational cost and GPU memory consumption, but achieves directional consistency.

To train our proposed method, we use prediction loss $L_{pred} $ and two auxillary losses $L_{aux}$. All losses are Ohem cross entropy loss \cite{shrivastava2016training}. $L_{aux, 1}$ is output features of the interpolated $P_{3}$. $L_{aux,2}$ is output features of the interpolated $P_{4}$. $\alpha, \beta$ are hyper parameters which balance $L_{total}$.
\begin{equation}
L_{total} = L_{pred} + \alpha L_{aux,1} + \beta L_{aux,2}
\end{equation}

\section{Experiments}
To compare the performance with other semantic segmentation approaches, we used Cityscapes dataset \cite{cordts2016cityscapes} which contains real driving environment. We used pre-trained ResNet18 with ImageNet \cite{krizhevsky2012imagenet} as the backbone. We trained the network with a batch size of 4, used SGD optimizer \cite{ruder2016overview} with a momentum of 0.9 and a weight decay of 0.0001. In training phase, we  used random color jittering, horizontal flipping, random scaling with 5 scales {0.75, 1.0, 1.5, 1.75, 2.0}. In testing phase, we resized input images of 1024 $\times$ 2048 to 1024 $\times$ 1024. For testing inference speed, we resized original input images to images of 512 $\times$ 1024 and we include resize operation in testing speed. Our method is tested on GTX 1080 Ti GPU with TensorRT.

\begin{table}[h]
\caption{Comparison of performance (mIoU, FPS) with other methods
on Cityscapes test datasets. * indicates using TensorRT framework.}
\resizebox{\linewidth}{!}{%
\begin{tabular}{cccc}
\hline
Approach     & Backbone    & Input Resolution & mIoU / FPS            \\ \hline
ICNet \cite{zhao2018icnet}        & ResNet50   & 1024$\times$2048 & 69.5 / 34             \\
LDFNet \cite{hung2019incorporating}& -   & 512$\times$1024        & 71.3 / 18.3           \\
DFANet A \cite{li2019dfanet}     & Xception A & 1024$\times$1024 & 71.3 / 100            \\
CellNet \cite{zhang2019customizable}     & -  & 768$\times$1536         & 70.5 / 108            \\
DF1-Seg* \cite{li2019partial}     & DF1  & 768$\times$1536       & 73 / 106.4            \\
FasterSeg* \cite{chen2019fasterseg}   & -   & 1024$\times$2048        & 71.5 / 163.9          \\
TinyHMSeg* \cite{li2020humans}  & -    & 768$\times$1536       & 71.4 / 172.4          \\
BiSeNetV2* \cite{yu2021bisenet}  & -    & 512$\times$1024       & 72.6 / 156            \\
STDC2-Seg50* \cite{fan2021rethinking} & STDC   & 512$\times$1024     & \textbf{73.4} / 188.6 \\ \hline
ASAP     & ResNet18  & 512$\times$1024  & 73.1 / 96  \\
ASAP*     & ResNet18  & 512$\times$1024  & 73.1 / \textbf{191}   \\ \hline
\end{tabular}%
}
\label{tab:miou}
\end{table}

\begin{table}[]
\centering
\caption{The effect of FFDN in combining multi level features. Performances are estimated on Cityscapes validation set.}
\begin{tabular}{cccc|c}
\hline
                              & *FPN & Layer Norm & Instance Norm & mIoU \\ \hline
\multicolumn{1}{c}{w/o FFDN} & \checkmark    & -                & -                   & 67.4 \\
\multicolumn{1}{c}{only LN}  &  \checkmark   &  \checkmark                & -                   & 70   \\
\multicolumn{1}{c}{only IN}  &  \checkmark   & -                &                \checkmark     & 74.9 \\
\multicolumn{1}{c}{FFDN}     &  \checkmark   &   \checkmark               &    \checkmark                 & \textbf{75.1} \\ \hline
\end{tabular}

\label{tab:ffdn}
\end{table}

\begin{table}[]
\caption{The comparison results of the feature fusion module and the self attention module on Cityscapes validation set. }
\centering

\begin{tabular}{cccc}
\hline
                                    &                & mIoU & GFLOPs \\ \hline
\multicolumn{1}{c|}{Feature Fusion} & General        & 74.4 & 1.08   \\
\multicolumn{1}{c|}{"}              & FFDN(ours)     & \textbf{75.9} & \textbf{0.54}   \\ \hline
\multicolumn{1}{c|}{Self Attention} & Conventional          & \textbf{76.7} & 87.52  \\
\multicolumn{1}{c|}{"}              & Horizontal     & 74.6 & 0.22   \\
\multicolumn{1}{c|}{"}              & Vertical(ours) & 75.9 & \textbf{0.22}   \\ \hline
\end{tabular}
\label{tab:flops}
\end{table}

\begin{figure*}[htb!]
\centering{\includegraphics[width=\linewidth]{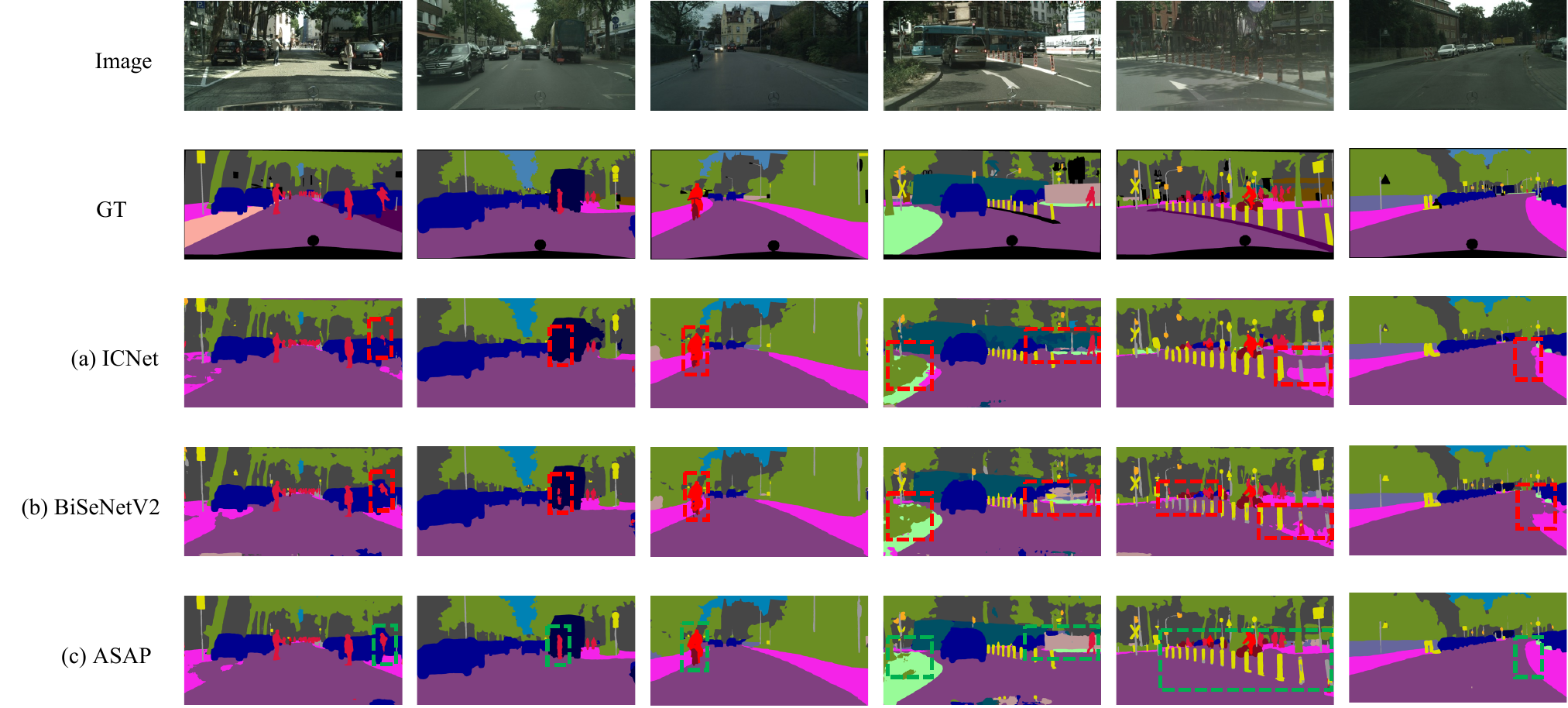}}
\caption{The visualized results of different methods including our proposed method on Cityscapes validation set. Red box represents challenging regions, and regions in green box are improved.}
\label{vis}
\end{figure*}

Table \ref{tab:miou} represents mIoU and FPS of experiment results on Cityscapes test dataset. As shown in Table \ref{tab:miou}, ASAPNet has mIoU of 73.1 and  FPS of 191. * indicates using TensorRT framework.

In terms of quantitive results, our method outperforms other real time targeted methods and has high performance in Table \ref{tab:miou}. 
Especially compared with STDC \cite{fan2021rethinking} which is the-state-of-the-arts, we achieve comparable performance. Although our method is about 0.3 mIoU below, we achieve about 3 frames fast. Therefore, we optimize both of fast inference speed and comparable performance. 

The reason is that our proposed method exploits instance-layer normalizations capturing essential spatial and semantic features instead of extracting global representations. Table \ref{tab:ffdn} shows the effect of our FFDN in combining multi level features. Since using only one normalization could handle only one property (e.g., spatial or semantic), it could not avoid to accuracy drops. 
In addition, the self-attention module with vertical pooling helps to keep directional consistency while requiring  computational complexity less than conventional self attention module.

Table \ref{tab:flops} shows that our two proposed module is optimized for both of accuracy and computational complexity. General means general feature fusion module extracting global perspective representations from multi level features by containing a convolution block(conv-batch norm-relu), a global average pooling, and a convolution block(conv-batch norm-sigmoid).
The general feature fusion requires huge computational complexity of 1.08G. However, our FFDN reduces about 2 times computations and increases accuracy by using different normalizations in combining multi level features. In the case of self attention, conventional self attention excluding a pooling operation achieves high accuracy but, it requires extremely computations of 87.52G. However, our self attention with vertical pooling could reduce computations about 400 times than the conventional self attention while the drops of accuracy is small of 0.8. In addition, we compare our vertical self attention with horizontal self attention which is processed after pooling horizontally. The vertical self attention outperforms the horizontal self attention by capturing challenging shallow objects. 

For a qualitive result, we provide qualitive comparison in shown in Fig \ref{vis}. A green box means the network segments the object correctly while a red box means the network fails to segment the object.

In first, second and third column, our self-attention with vertical axis helps to focus on shallow objects (e.g., person, pole and bicycle) while other methods loss them.  In addition, we find out that the attention helps the network to focus on each object. In fourth, fifth and last column, our method could discriminate huge objects (e.g., road, wall, vegetation, terrain and sidewalk) by using our FFDN capturing spatial and semantic representations while ICNet based on only convolution layers could not discriminate them.
Therefore, our method takes advantage of Advanced Driver Assistance System(ADAS) to interpret the driving path, such as calculating the curvature of a curved road for path planning.


\section{Conclusion}
In this paper, we analyze the low performance of specific objects, and propose novel self-attention module using vertical axis pooling. The complexity of the module is much lower than state-of-the-art. Also, we propose the feature fusion module, exploiting different normalization techniques. Using both of two module, we save computational costs while we achieve high performance in Cityscapes test datasets.

\bibliographystyle{IEEEtran}
\bibliography{main}

\end{document}